\pdfoutput=1

\documentclass[11pt]{article}

\usepackage[]{naacl2021}

\usepackage{times}
\usepackage{amsmath}
\usepackage{amssymb}
\usepackage{graphicx}
\usepackage{caption}
\usepackage{subcaption}
\usepackage{float}
\usepackage{booktabs}
\usepackage{natbib}
\usepackage{latexsym}
\usepackage{todonotes}
\usepackage{algorithmicx}
\usepackage[ruled,vlined]{algorithm2e}

\usepackage[T1]{fontenc}

\usepackage[utf8]{inputenc}

\usepackage{microtype}

\title{Robust Question Answering Through Sub-part Alignment}

\author{Jifan Chen \and Greg Durrett \\
  The University of Texas at Austin \\
  {\tt \{jfchen, gdurrett\}@cs.utexas.edu}}

\date{}

\begin{document}
\maketitle
\begin{abstract}
Current textual question answering (QA) models achieve strong performance on in-domain test sets, but often do so by fitting surface-level patterns, so they fail to generalize to out-of-distribution settings. To make a more robust and understandable QA system, we model question answering as an alignment problem. We decompose both the question and context into smaller units based on off-the-shelf semantic representations (here, semantic roles), and align the question to a subgraph of the context in order to find the answer. We formulate our model as a structured SVM, with alignment scores computed via BERT, and we can train end-to-end despite using beam search for approximate inference. Our use of explicit alignments allows us to explore a set of constraints with which we can prohibit certain types of bad model behavior arising in cross-domain settings. Furthermore, by investigating differences in scores across different potential answers, we can seek to understand what particular aspects of the input lead the model to choose the answer without relying on post-hoc explanation techniques. We train our model on SQuAD v1.1 and test it on several adversarial and out-of-domain datasets. The results show that our model is more robust than the standard BERT QA model, and constraints derived from alignment scores allow us to effectively trade off coverage and accuracy.

\end{abstract}

\section{Introduction}
Current text-based question answering models learned end-to-end often rely on spurious patterns between the question and context rather than learning the desired behavior. They may ignore the question entirely \cite{kaushik2018much}, focus primarily on the answer type \cite{mudrakarta2018did}, or otherwise bypass the ``intended'' mode of reasoning for the task \cite{chen2019understanding,niven2019probing}. Thus, these models are not robust to adversarial attacks \cite{jia2017adversarial, iyyer2018adversarial, Wallace2019Triggers}: they can be fooled by surface-level distractor answers that follow the spurious patterns. Methods like adversarial training~\cite{miyato2016adversarial, wang2018robust, lee2019domain, yang2019improving}, data augmentation~\cite{welbl2020undersensitivity}, and posterior regularization~\cite{pereyra2016regularizing, zhou2019robust} have been proposed to improve robustness. However, these techniques often optimize for a certain type of error. We want models that can adapt to new types of adversarial examples and work under other distribution shifts, such as on questions from different text domains \cite{fisch2019mrqa}.

\begin{figure}[t]
\centering
\includegraphics[width=0.48\textwidth]{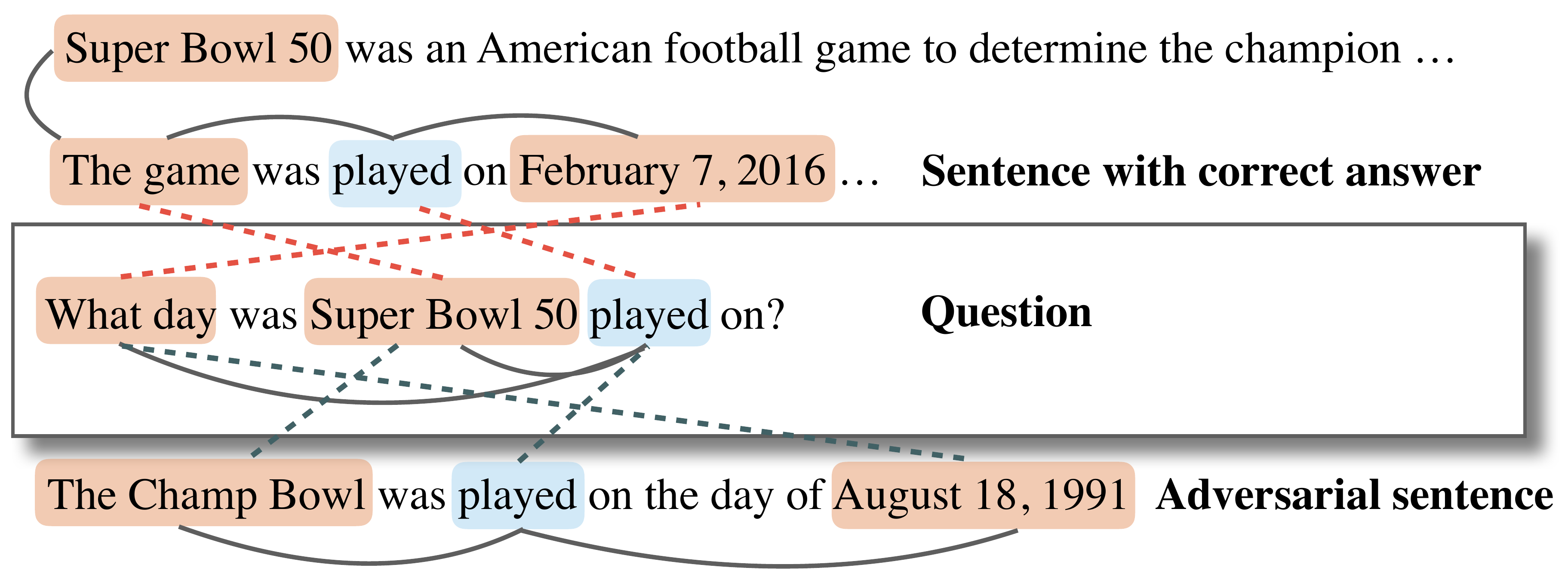}
\vspace{-0.5cm}
\caption{A typical example on adversarial SQuAD. By breaking the question and context down into smaller units, we can expose the incorrect entity match and use explicit constraints to fix it. The solid lines denote edges from SRL and coreference, and the dotted lines denote the possible alignments between the arguments (desired in red, actual in black).}
\vspace{-0.3cm}
    \label{fig:intro_example}
\end{figure}

In this paper, we explore a model for text-based question answering through sub-part alignment. The core idea behind our method is that if every aspect of the question is well supported by the answer context, then the answer produced should be trustable~\cite{lewis2018generative}; if not, we suspect that the model is making an incorrect prediction. The sub-parts we use are predicates and arguments from Semantic Role Labeling \cite{palmer2005srl}, which we found to be a good semantic representation for the types of questions we studied. We then view the question answering procedure as a constrained graph alignment problem \cite{sachan2016machine}, where the nodes represent the predicates and arguments and the edges are formed by relations between them (e.g. predicate-argument relations and coreference relations). Our goal is to align each node in the question to a counterpart in the context, respecting some loose constraints, and in the end the context node aligned to the wh-span should ideally contain the answer. Then we can use a standard QA model to extract the answer.

Figure~\ref{fig:intro_example} shows an adversarial example of SQuAD~\cite{jia2017adversarial} where a standard BERT QA model predicts the wrong answer \emph{August 18, 1991}. In order to choose the adversarial answer, our model must \textbf{explicitly} align \emph{Super Bowl 50} to \emph{Champ Bowl}. Even if the model still makes this mistake, this error is now exposed directly, making it easier to interpret and subsequently patch.


In our alignment model, each pair of aligned nodes is scored using BERT \cite{devlin2019bert}. These alignment scores are then plugged into a beam search inference procedure to perform the constrained graph alignment. This structured alignment model can be trained as a structured support vector machine (SSVM) to minimize alignment error with heuristically-derived oracle alignments. The alignment scores are computed in a black-box way, so these individual decisions aren't easily explainable~\cite{jain2019attention}; however, the score of an answer is directly a sum of the score of each aligned piece, making this structured prediction phase of the model faithful by construction~\cite{jain2020learning}. Critically, this allows us to understand what parts of the alignment are responsible for a prediction, and if needed, constrain the behavior of the alignment to correct certain types of errors. We view this interpretability and extensibility with constraints as one of the principal advantages of our model.

We train our model on the SQuAD-1.1 dataset~\cite{rajpurkar2016squad} and evaluate on SQuAD Adversarial~\cite{jia2017adversarial}, Universal Triggers on SQuAD~\cite{Wallace2019Triggers}, and several out-of-domain datasets from MRQA~\cite{fisch2019mrqa}. Our framework allows us to incorporate natural constraints on alignment scores to improve zero-shot performance under these distribution shifts, as well as explore coverage-accuracy tradeoffs in these settings. Finally, our model's alignments serve as ``explanations'' for its prediction, allowing us to ask why certain predictions are made over others and examine scores for hypothetical other answers the model could give.

\begin{figure}[t]
\centering
\includegraphics[width=0.48\textwidth]{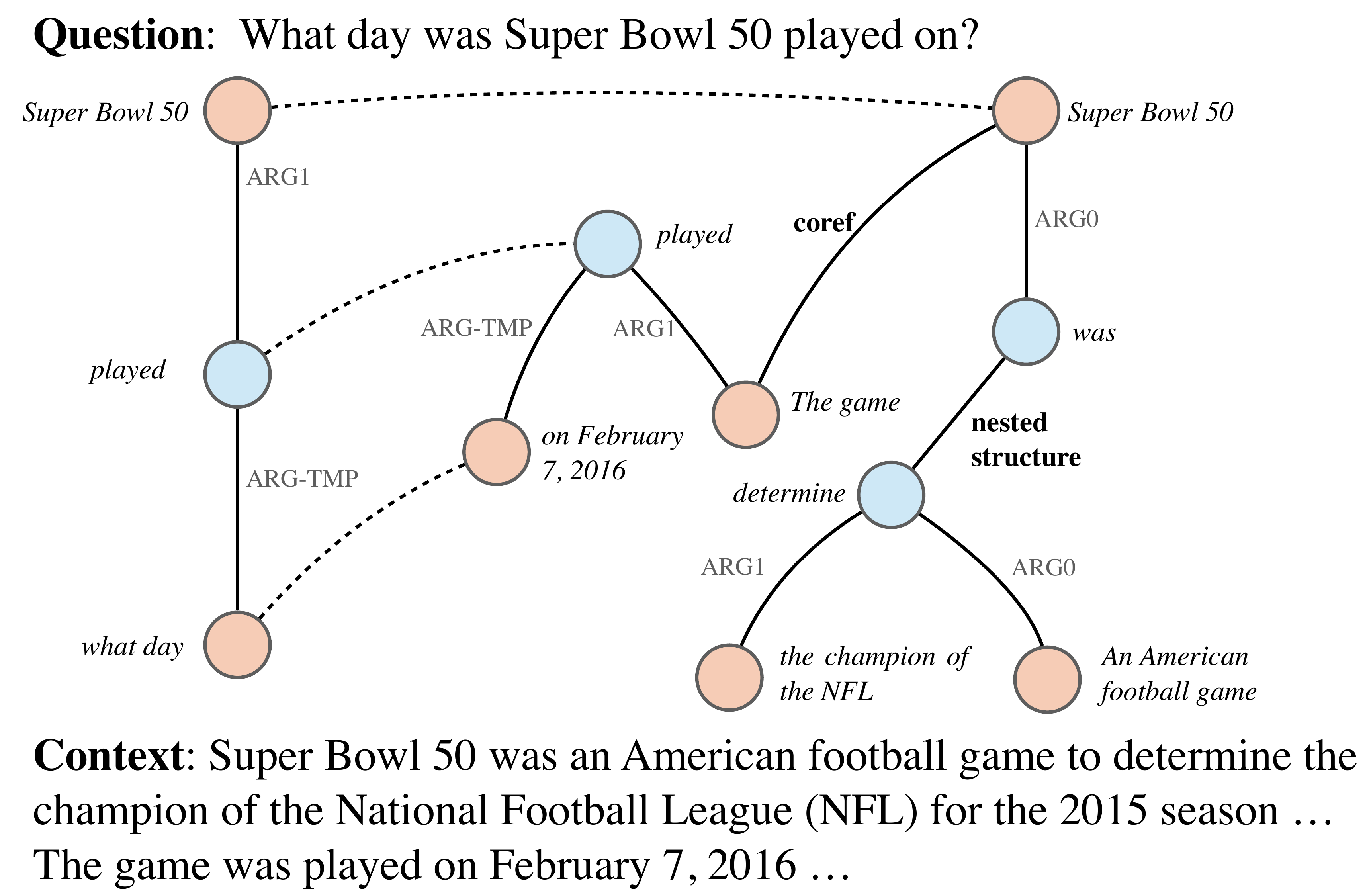}
\vspace{-0.5cm}
\caption{Example of our question-passage graph. Edges come from SRL, coreference (\emph{Super Bowl 50}---\emph{the game}), and postprocessing of predicates nested inside arguments (\emph{was}---\emph{determine}). The oracle alignment (Section~\ref{sec:oracle}) is shown with dotted lines. Blue nodes are predicates and orange ones are arguments.}

\vspace{-0.2cm}
    \label{fig:graph_alignment}
\end{figure}

\section{QA as Graph Alignment}

Our approach critically relies on the ability to decompose questions and answers into a graph over text spans. Our model can in principle work for a range of syntactic and semantic structures, including dependency parsing, SRL \cite{palmer2005srl}, and AMR~\cite{banarescu2013abstract}. We use SRL in this work and augment it with coreference links, due to the high performance and flexibility of current SRL systems~\cite{peters2018deep}. Throughout this work, we use the BERT-based SRL system from \citet{shi2019simple} and the SpanBERT-based coreference system from \citet{joshi2020spanbert}.

An example graph we construct is shown in Figure~\ref{fig:graph_alignment}. Both the question and context are represented as graphs where the nodes consist of predicates and arguments. Edges are undirected and connect each predicate and its corresponding arguments. Since SRL only captures the predicate-argument relations within one sentence, we add coreference edges as well: if two arguments are in the same coreference cluster, we add an edge between them. Finally, in certain cases involving verbal or clausal arguments, there might exist nested structures where an argument to one predicate contains a separate predicate-argument structure. In this case, we remove the larger argument and add an edge directly between the two predicates. This is shown by the edge from \emph{was} to \emph{determine} (labeled as \emph{nested structure}) in Figure~\ref{fig:graph_alignment}). Breaking down such large arguments helps avoid ambiguity during alignment. 

Aligning questions and contexts has proven useful for question answering in previous work~\cite{sachan2015learning, sachan2016machine,khashabi2018question}. Our framework differs from theirs in that it incorporates a much stronger alignment model (BERT), allowing us to relax the alignment constraints and build a more flexible, higher-coverage model.


\paragraph{Alignment Constraints} Once we have the constructed graph, we can align each node in the question to its counterpart in the context graph. In this work, we control the alignment behavior by placing explicit constraints on this process. We place a \textbf{locality constraint} on the alignment: adjacent pairs of question nodes must align no more than $k$ nodes apart in the context. $k=1$ means we are aligning the question to a connected sub-graph in the context, $k=\infty $ means we can align to a node anywhere in a connected component in the context graph. In our experiments, we set $k=3$. In the following sections, we will discuss more constraints. Altogether, these constraints define a set $\mathcal{A}$ of possible alignments.

\section{Graph Alignment Model}

\subsection{Model}

Let $\mathbf{T}$ represent the text of the context and question concatenated together. Assume a decomposed question graph $\mathbf{Q}$ with nodes $q_1, q_2, \ldots , q_m$ represented by vectors $\mathbf{q}_1, \mathbf{q}_2, \ldots , \mathbf{q}_m$, and a decomposed context $\mathbf{C}$ with nodes $c_1,\ldots, c_n$ represented by vectors $\mathbf{c_1},\ldots, \mathbf{c}_n$. Let $\mathbf{a} = (a_1,\ldots,a_m)$ be an alignment of question nodes to context nodes, where $a_i \in \{1,\ldots,n\}$ indicates the alignment of the $i$th question node. Each question node is aligned to exactly one context node, and multiple question nodes can align to the same context node.

We frame question answering as a maximization of an alignment scoring function over possible alignments:
$\max_{\mathbf{a} \in \mathcal{A}} f(\mathbf{a}, \mathbf{Q}, \mathbf{C}, \mathbf{T})$. In this paper, we simply choose $f$ to be the sum over the scores of all alignment pairs $f(\mathbf{a}, \mathbf{Q}, \mathbf{C}, \mathbf{T}) = \sum_{i=1}^n S(q_i, c_{a_i}, \mathbf{T})$, where $S(q,c, \mathbf{T})$ denotes the alignment score between a question node $q$ and a context node $c$. This function relies on BERT \cite{devlin2019bert} to compute embeddings of the question and context nodes and will be described more precisely in what follows. We will train this model as a structured support vector machine (SSVM), described in Section~\ref{sec:training}.

\begin{figure}[t]
\centering
\includegraphics[width=0.48\textwidth]{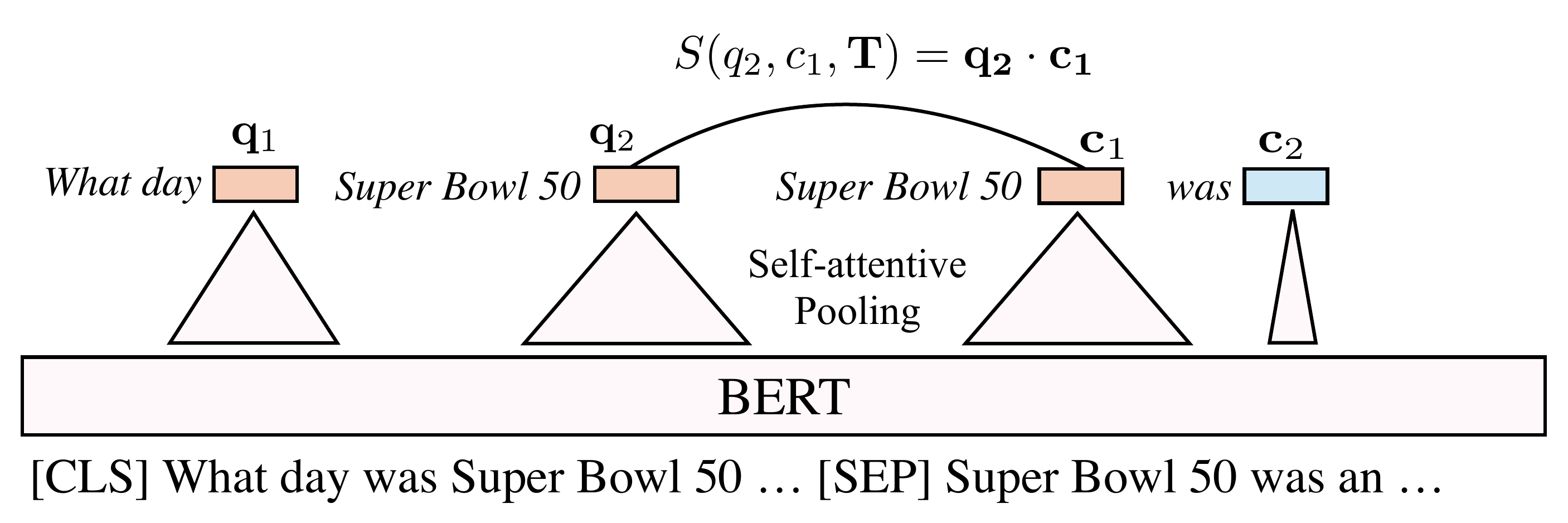}
\caption{Alignment scoring. Here the alignment score is computed by the dot product between span representations of question and context nodes. The final alignment score (not shown) is the sum of these edge scores.}
    \label{fig:model_ssvm}
\end{figure}

\paragraph{Scoring} Our alignment scoring process is shown in Figure~\ref{fig:model_ssvm}. We first concatenate the question text with the document text into $\mathbf{T}$ and then encode them using the pre-trained BERT encoder. We then compute a representation for each node in the question and context using a span extractor, which in our case is the self-attentive pooling layer of~\citet{lee2017end}. The node representation in the question can be computed in the same way. Then the score of a node pair is computed as a dot product $S(q, c,\mathbf{T}) = \mathbf{q} \cdot \mathbf{c}$.  


\paragraph{Answer Extraction} Our model so far produces an alignment between question nodes and context nodes. We assume that one question node contains a wh-word and this node aligns to the context node containing the answer.\footnote{We discuss what to do with other questions in Section~\ref{sec:experimental_setup}.} Ideally, we can use this aligned node to extract the actual answer. However, in practice, the aligned context node may only contain part of the answer and in some cases answering the question only based the aligned context node can be ambiguous. We therefore use the sentence containing the wh-aligned context node as the ``new'' context and use a standard BERT QA model to extract the actual answer post-hoc. In the experiments, we also show the performance of our model by only use the aligned context node without the sentence, which is only slightly worse.


\subsection{Training}
\label{sec:training}

We train our model as an instance of a structured support vector machine (SSVM). Ignoring the regularization term, this objective can be viewed as a sum over the training data of a structured hinge loss with the following formulation:\vspace{-0.3cm}
\begin{equation*}
    \begin{split}
            \sum_{i=1}^N \max(0, \max_{\mathbf{a} \in \mathcal{A}} [ &f(\mathbf{a}, \mathbf{Q}_i, \mathbf{C}_i, \mathbf{T}_i) + \text{Ham}(\mathbf{a}, \mathbf{a}_i^*)] \\ & - f(\mathbf{a}_i^*, \mathbf{Q}_i, \mathbf{C}_i, \mathbf{T}_i) ])
    \end{split}
\end{equation*}
where $\mathbf{a}$ denotes the predicted alignment, $\mathbf{a}_i^*$ is the oracle alignment for the $i$th training example, and $\text{Ham}$ is the Hamming loss between these two.  To get the predicted alignment $\mathbf{a}$ during training, we need to run loss-augmented inference as we will discuss in the next section. When computing the alignment for node $j$, if $a_{j} \neq a_j^*$, we add 1 to the alignment score to account for the loss term in the above equation. Intuitively, this objective requires the score of the gold prediction to be larger than any other hypothesis $\mathbf{a}$ by a margin of  $\text{Ham}(\mathbf{a}, \mathbf{a}^*)$. 

When training our system, we first do several iterations of \emph{local training} where we treat each alignment decision as an independent prediction, imposing no constraints, and optimize log loss over this set of independent decisions. The local training helps the global training converge more quickly and achieve better performance.

\subsection{Inference}
\label{sec:inference}

Since our alignment constraints do not strongly restrict the space of possible alignments (e.g., by enforcing a one-to-one alignment with a connected subgraph), searching over all valid alignments is intractable. We therefore use beam search to find the approximate highest-scoring alignment: (1) Initialize the beam with top $b$ highest aligned node pairs, where $b$ is the beam size. (2) For each hypothesis (partial alignment) in the beam, compute a set of reachable nodes based on the currently aligned pairs under the locality constraint. (3) Extend the current hypothesis by adding each of these possible alignments in turn and accumulating its score. Beam search continues until all the nodes in the question are aligned.

\begin{figure}[t]
\centering
\includegraphics[width=0.48\textwidth]{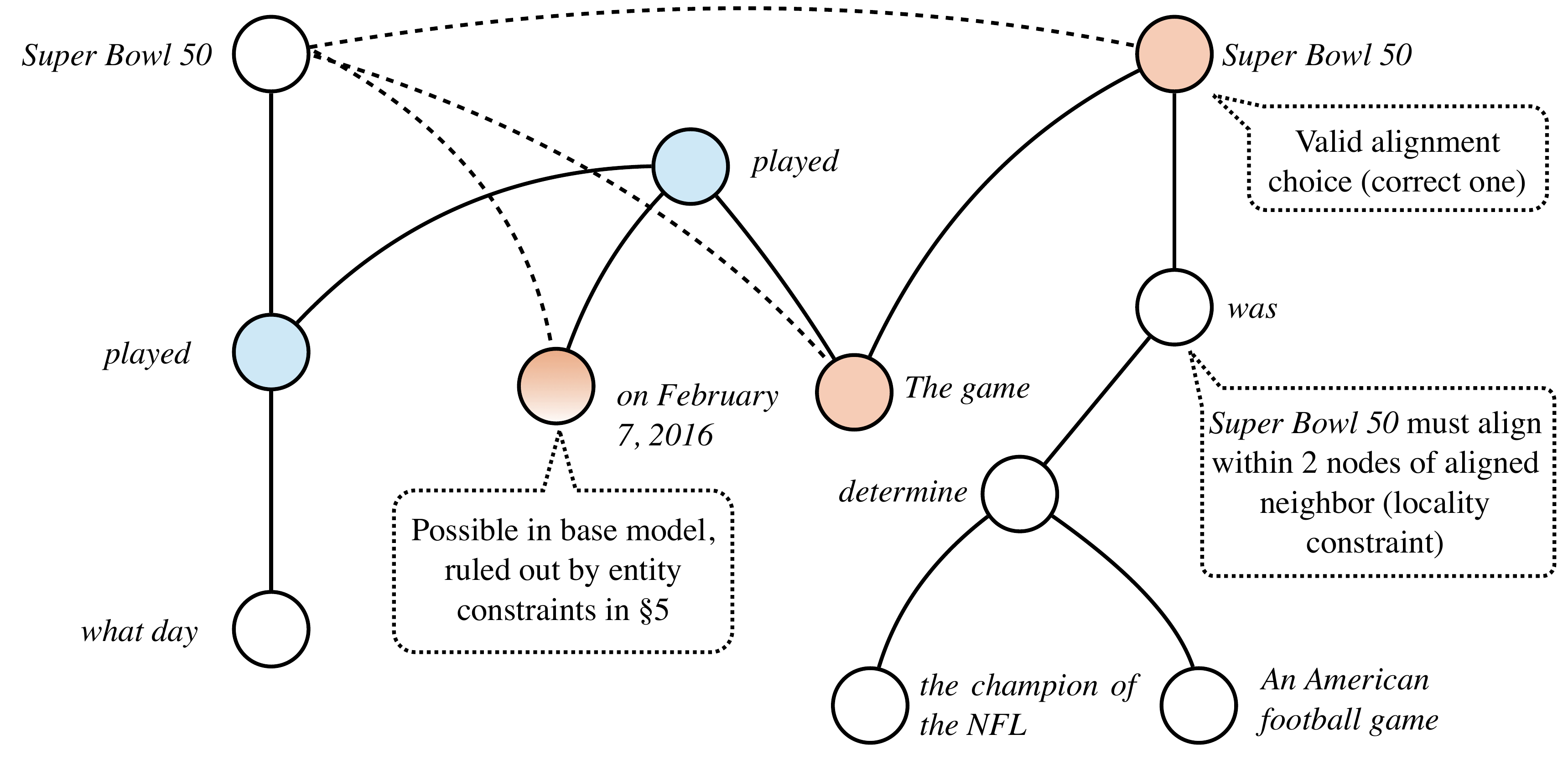}
\caption{An example of constraints during beam search. The blue node \emph{played} is already aligned. The orange nodes denote all the valid context nodes that can be aligned to for both \emph{Super Bowl 50} and \emph{what day} in the next step of inference given the locality constraint with $k=2$.}
\vspace{-0.5cm}
    \label{fig:beam_search}
\end{figure}

An example of one step of beam hypothesis expansion with locality constraint $k=2$ is shown in Figure~\ref{fig:beam_search}. In this state, the two \emph{played} nodes are already aligned. In any valid alignment, the neighbors of the \emph{played} question node must be aligned within 2 nodes of the \emph{played} context node to respect the locality constraint. We therefore only consider aligning to \emph{the game}, \emph{on Feb 7, 2016} and \emph{Super Bowl 50}. The alignment scores between these reachable nodes and the remaining nodes in the question are computed and used to extend the beam hypotheses. 

Note that this inference procedure allows us to easily incorporate other constraints as well. For instance, we could require a ``hard'' match on entity nodes, meaning that two nodes containing entities can only align if they share entities. With this constraint, as shown in the figure, \emph{Super Bowl 50} can never be aligned to \emph{on February 7, 2016}. We discuss such constraints more in Section~\ref{sec:constraints}.

\subsection{Oracle Construction}
\label{sec:oracle}
Training assumes the existence of gold alignments $\mathbf{a}^*$, which must be constructed via an oracle given the ground truth answer. This process involves running inference based on heuristically computed alignment scores $S_{\text{oracle}}$, where $S_{\text{oracle}}(q,c)$ is computed by the Jaccard similarity between a question node $q$ and a context node $c$. Instead of initializing the beam with the $b$ best alignment pairs, we first align the wh-argument in the question with the node(s) containing the answer in the context and then initialize the beam with those alignment pairs.

If the Jaccard similarity between a question node and all other context nodes is zero, we set these as unaligned nodes. 
During training, our approach can gracefully handle unaligned nodes by treating these as latent variables in structured SVM: the gold ``target'' is then highest scoring set of alignments consistent with the gold supervision. This involves running a second decoding step on each example to impute the values of these latent variables for the gold alignment. 

\begin{table*}[t]
\small
\centering
\renewcommand{\tabcolsep}{0.9mm}
\begin{tabular}{ l | c c | c c | c c |  c c | c c | c c }
\toprule
            & \multicolumn{2}{c|}{SQuAD normal} & \multicolumn{2}{c|}{SQuAD addSent} &  \multicolumn{2}{c|}{Natural Questions} & \multicolumn{2}{c|}{NewsQA} & \multicolumn{2}{c|}{BioASQ} & \multicolumn{2}{c}{TBQA} \\
\midrule
            & ans in wh & F1 & ans in wh & F1 & ans in wh & F1 & ans in wh & F1 & ans in wh & F1 & ans in wh & F1 \\
\midrule
Sub-part Alignment & 84.7 & 84.5 & 49.5 & \textbf{50.5}& 65.8 & 61.5 & 49.3 & 48.1 & 63.5 & \textbf{53.4} & 35.1 & \textbf{38.4} \\
\midrule
$-$ global train+inf  & 85.8  & 85.2  & 45.0  & 46.8 & 65.9  & \textbf{62.3}  & 48.9  & 47.1 & 62.5 & 52.1 & 31.9 & 34.6 \\
$-$ ans from full sent & 84.7  & 81.8  & 49.5  & 46.7 & 65.8  & 57.8  & 49.3  & 45.0 & 63.5 & 51.1 & 35.1 & 37.5 \\
\midrule
BERT QA & $-$ & \textbf{87.8} & $-$ & 39.2 & $-$ & 59.4 & $-$ & \textbf{48.5} & $-$ & 52.4 & $-$ & 25.3 \\
\bottomrule
\end{tabular}
\caption{The performance and ablations of our proposed model on the development sets of SQuAD, adversarial SQuAD, and four out-of-domain datasets. Our Sub-part Alignment model uses both global training and inference as discussed in Section~\ref{sec:training}-\ref{sec:inference}. $-$ \textbf{global train+inf} denotes the locally trained and evaluated model. $-$ \textbf{ans from full sent} denotes extracting the answer using only the wh-aligned node. \textbf{ans in wh} denotes the percentage of answers found in the span aligned to the wh-span, and F1 denotes the standard QA performance measure. Here for \texttt{\small addSent}, we only consider the adversarial examples. Note also that this evaluation is \emph{only on wh-questions}.}
\vspace{-0.1cm}
\label{tab:res_squad_adv}
\end{table*}

\section{Experiments: Adversarial and Cross-domain Robustness}

Our focus in this work is primarily robustness, interpretability, and controllability of our model. We focus on adapting to challenging settings in order to ``stress test'' our approach.

\subsection{Experimental Settings}
\label{sec:experimental_setup}
For all experiments, we train our model \emph{only} on the English SQuAD-1.1 dataset \cite{rajpurkar2016squad} and examine how well it can generalize to adversarial and out-of-domain settings with minimal modification, using \emph{no fine-tuning} on new data and \emph{no data augmentation} that would capture useful transformations. We evaluate on the \texttt{addSent} and \texttt{addOneSent} proposed by~\newcite{jia2017adversarial}, and the \texttt{Universal Triggers on SQuAD}~\cite{Wallace2019Triggers}.  We also test the performance of our SQuAD-trained models in zero-shot adaptation to new English domains, namely Natural Questions~\cite{kwiatkowski2019natural}, NewsQA~\cite{trischler2017newsqa}, BioASQ~\cite{bioasq2015} and TextbookQA~\cite{kembhavi2017you}, taken from the MRQA shared task~\cite{fisch2019mrqa}. Our motivation here was to focus on text from a variety of domains where transferred SQuAD models may at least behave credibly. We excluded, for example, HotpotQA~\cite{yang2018hotpotqa} and DROP~\cite{dua2019drop}, since these are so far out-of-domain from the perspective of SQuAD that we do not see them as a realistic cross-domain target. 

We compare primarily against a standard \textbf{BERT QA} system \cite{devlin2019bert}. We also investigate a local version of our model, where we only try to align each node in the question to its oracle, without any global training (\textbf{$-$ global train + inf}), which can still perform reasonably because BERT embeds the whole question and context. When comparing variants of our proposed model, we only consider the questions that have a valid SRL parse and have a wh word (results in Table~\ref{tab:res_squad_adv}, Table~\ref{tab:res_squad_adv_trigger}, and Figure~\ref{fig:f1_coverage_curve}). When comparing with prior systems, for questions that do not have a valid SRL parse or wh word, we back off to the standard BERT QA system (results in Table~\ref{tab:res_other_systems}).

We set the beam size $b=20$ for the constrained alignment. We use \texttt{BERT-base-uncased} for all of our experiments, and fine-tune the model using Adam~\cite{kingma2014adam} with learning rate set to 2e-5. Our preprocessing uses a SpanBERT-based coreference system~\cite{joshi2020spanbert} and a BERT-based SRL system~\cite{shi2019simple}. We limit the length of the context to 512 tokens. For our global model, we initialize the weights using a locally trained model and then fine-tune using the SSVM loss. We find the initialization helps the model converge much faster and it achieves better performance than learning from scratch. When doing inference, we set the locality constraint $k=3$.




\subsection{Results on Challenging Settings}

The results\footnote{Here we omit \texttt{SQuAD addOneSent} for simplicity, since the performance on it has the same trend as \texttt{SQuAD addSent}. Refer to the Appendix for the results on \texttt{SQuAD addOneSent}.} on the normal SQuAD development set and other challenging sets are shown in Table~\ref{tab:res_squad_adv}.

\paragraph{Our model is not as good as BERT QA on normal SQuAD but outperforms it in challenging settings.} Compared to the BERT QA model, our model is fitting a different data distribution (learning a constrained structure) which makes the task harder. This kind of training scheme does cause some performance drop on normal SQuAD, but we can see that it consistently improves the F1 on the adversarial (on SQuAD addSent, a 11.3 F1 improvement over BERT QA) and cross-domain datasets except NewsQA (where it is 0.4 F1 worse). This demonstrates that learning the alignment helps improve the robustness of our model.

\paragraph{Global training and inference improve performance in adversarial settings, despite having no effect in-domain.} Normal SQuAD is a relatively easy dataset and the answer for most questions can be found by simple lexical matching between the question and context. From the ablation of $-$ \textbf{global train+inf}, we can see that more than 80\% of answers can be located by matching the wh-argument. We also observe a similar pattern on Natural Questions.\footnote{For the MRQA task, only the paragraph containing the short answer of NQ is provided as context, which eliminates many distractors. In such cases, those NQ questions have a similar distribution as those in SQuAD-1.1, and similarly make no use of the global alignment.} However, as there are very strong distractors in \texttt{SQuAD addSent}, the wh-argument matching is unreliable. In such situations, the constraints imposed by other argument alignments in the question are useful to correct the wrong wh-alignment through global inference. We see that the global training plus inference is consistently better than the local version on all other datasets.


\paragraph{Using the strict wh answer extraction still gives strong performance} From the ablation of \textbf{$-$ ans from full sent}, we observe that our ``strictest'' system that extracts the answer only using the wh-aligned node is only worse by 3-4 points of F1 on most datasets. Using the full sentence gives the system more context and maximal flexibility, and allows it to go beyond the argument spans introduced by SRL. We believe that better semantic representations tailored for question answering~\cite{lamm2020qed} will help further improvement in this regard.

\begin{table}[t]
\small
\centering
\begin{tabular}{ c | c p{0.5cm}  c |  c p{0.5cm}  c }
\toprule
& \multicolumn{3}{|c}{Sub-part Alignment} & \multicolumn{3}{|c}{BERT}\\
    Type        & Normal & Trigger & $\Delta$  & Normal & Trigger & $\Delta$   \\
\midrule
who & 84.7 & 82.7 & 2.0 & 87.1 & 78.5 & 8.6 \\
why & 75.1 & 71.3 & 3.8 & 76.5 & 59.7 & 16.8 \\
when & 88.4 & 82.8 & 5.6 & 90.3 & 80.9 & 9.4 \\
where & 83.6  & 81.4 & 2.2 & 84.1 & 75.8 & 8.3 \\


\bottomrule
\end{tabular}
\caption{The performance of our model on the Universal Triggers on SQuAD dataset \cite{Wallace2019Triggers}. Compared with BERT, our model sees smaller performance drops on all triggers.}
\vspace{-0.2cm}
\label{tab:res_squad_adv_trigger}
\end{table}

\begin{table*}[t]
\small
\centering
\begin{tabular}{ c | c | c c c | c c c }
\toprule
            & Normal & \multicolumn{3}{c|}{addSent}& \multicolumn{3}{c}{addOneSent} \\
\midrule
            &  & overall & adv & $\Delta$ & overall & adv & $\Delta$\\
\midrule
R.M-Reader~\cite{hu2018reinforced}  & 86.6  & 58.5  & $-$ & 31.1 & 67.0 & $-$ & 19.6 \\  
KAR~\cite{wang2018explicit}         & 83.5 & 60.1   & $-$ & 23.4  & 72.3 & $-$ & \textbf{11.2} \\
BERT + Adv~\cite{yang2019improving}  & 92.4  & 63.5 &  $-$ & 28.9 & 72.5 & $-$ & 19.9 \\

\midrule
Our BERT & 87.8  & 61.8  & 39.2 & 27.0  & 70.4  & 52.6 & 18.4 \\
Sub-part Alignment* & 84.7 & \textbf{65.8}  & 47.1 & \textbf{18.9} & \textbf{72.8} & 60.1 & 11.9 \\

\bottomrule
\end{tabular}
\caption{ Performance of our systems compared to the literature on both \texttt{addSent} and \texttt{addOneSent}. Here, overall denotes the performance on the full adversarial set, adv denotes the performance on the adversarial samples alone. $\Delta$ represents the gap between the normal SQuAD and the overall performance on adversarial set.}
\vspace{-0.2cm}
\label{tab:res_other_systems}
\end{table*}

\subsection{Results on Universal Triggers}
The results on subsets of the universal triggers dataset are shown in Table~\ref{tab:res_squad_adv_trigger}. We see that every trigger  results in a bigger performance drop on BERT QA than our model. Our model is much more stable, especially on \emph{who} and \emph{where} question types, in which case the performance only drops by around 2\%. Several factors may contribute to the stability: (1) The triggers are ungrammatical and their arguments often contain seemingly random words, which are likely to get lower alignment scores. (2) Because our model is structured and trained to align all parts of the question, adversarial attacks on span-based question answering models may not fool our model as effectively as they do BERT.

\subsection{Comparison to Existing Systems}

In Table~\ref{tab:res_other_systems}, we compare our best model (not using constraints from Section~\ref{sec:constraints}) with existing adversarial QA models in the literature. We note that the performance of our model on SQuAD-1.1 data is relatively lower compared to those methods, yet we achieve the best overall performance; we trade some in-distribution performance to improve the model's robustness. We also see that our model achieves the smallest normal vs.~adversarial gap on \texttt{addSent} and  \texttt{addOneSent}, which demonstrates that our constrained alignment process can enhance the robustness of the model compared to prior methods like adversarial training~\cite{yang2019improving} or explicit knowledge integration~\cite{wang2018explicit}.

\section{Generalizing by Alignment Constraints}
\label{sec:constraints}

One advantage of our explicit alignments is that we can understand and inspect the model's behavior more deeply. This structure also allows us to add constraints to our model to prohibit certain behaviors, which can be used to adapt our model to adversarial settings.

In this section, we explore how two types of constraints enable us to reject examples the model is less confident about. Hard constraints can enable us to reject questions where the model finds no admissible answers. Soft constraints allow us to set a calibration threshold for when to return our answer. We focus on evaluating our model's accuracy at various coverage points, the so-called selective question answering setting~\cite{kamath2020selective}. 


\paragraph{Constraints on Entity Matches}
By examining \texttt{addSent} and \texttt{addOneSent}, we find the model is typically fooled when the nodes containing entities in the question align to ``adversarial'' entity nodes. An intuitive constraint we can place on the alignment is that we require a hard entity match---for each argument in the question, if it contains entities, it can only align to nodes in the context sharing exact the same entities.

\paragraph{Constraints on Alignment Scores}
The hard entity constraint is quite inflexible and does not generalize well, for example to questions that do not contain a entity. However, the alignment scores we get during inference time are good indicators of how well a specific node pair is aligned. For a correct alignment, every pair should get a reasonable alignment score. However, if an alignment is incorrect, there should exist some bad alignment pairs which have lower scores than the others. We can reject those samples by finding bad alignment pairs, which both improves the precision of our model and also serves as a kind of explanation as to why our model makes its predictions.

We propose to use a simple heuristic to identify the bad alignment pairs. We first find the max score $S_{\max}$ over all possible alignment pairs for a sample, then for each alignment pair $(q_i, c_j)$ of the prediction, we calculate the worst alignment gap (WAG) $g = \text{min}_{(q,c) \in \mathbf{a}} (S_{\max} - S(q, c)) $. If $g$ is beyond some threshold, it indicates that alignment pair is not reliable.\footnote{The reason we look at differences from the max alignment is to calibrate the scores based on what ``typical'' scores look like for that instance. We find that these are on different scales across different instances, so the gap is more useful than an absolute threshold.}


\begin{figure}[t]
\centering
\includegraphics[width=0.5\textwidth]{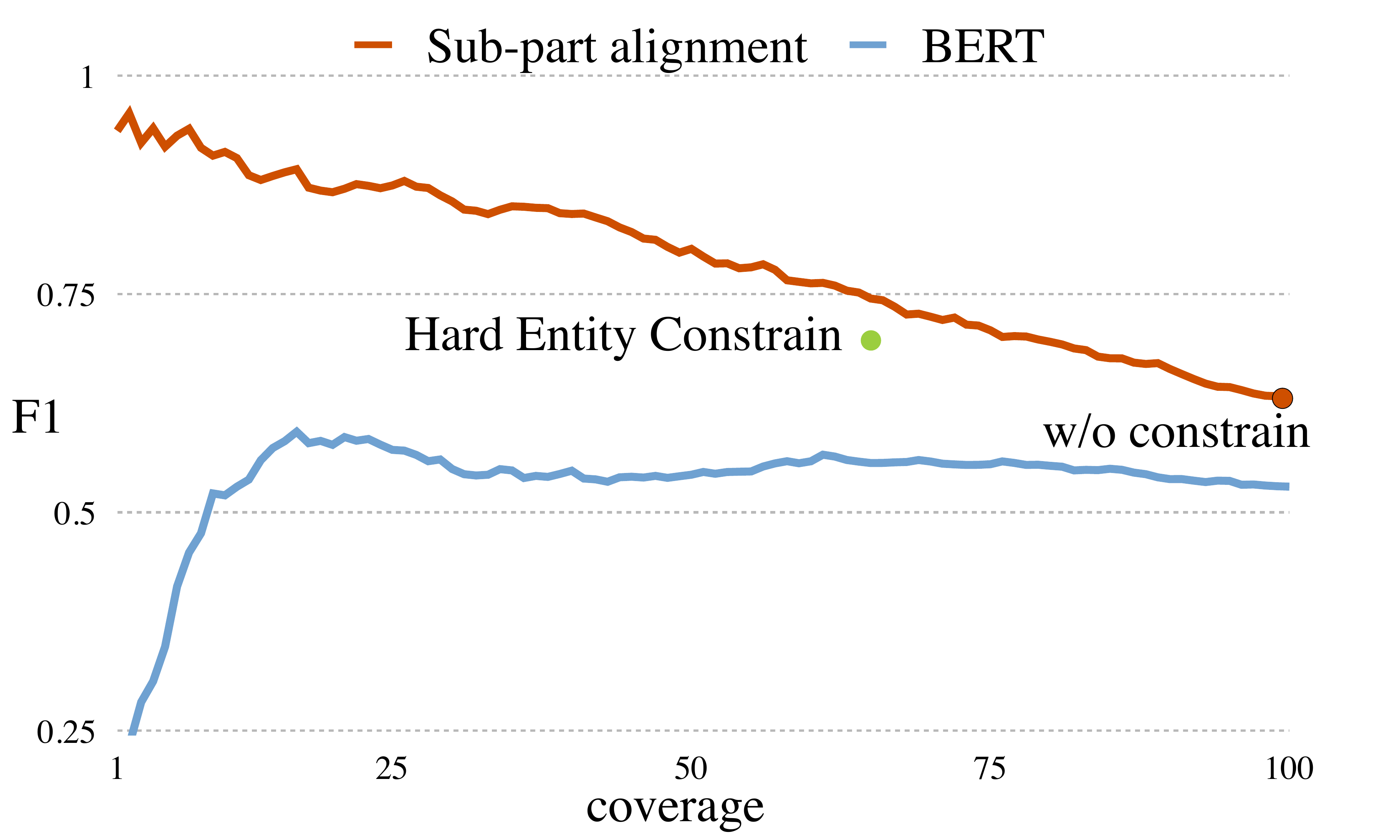}
\vspace{-0.2cm}
\caption{The F1-coverage curve of our model compared with BERT QA. If our model can choose to answer only the $k$ percentage of examples it's most confident about (the coverage), what F1 does it achieve? For our model, the confidence is represented by our ``worst alignment gap'' (WAG) metric. Smaller WAG indicates higher confidence. For BERT, the confidence is represented by the posterior probability.}
\vspace{-0.2cm}
    \label{fig:f1_coverage_curve}
\end{figure}

\begin{figure*}[t]
\centering
\includegraphics[width=\textwidth]{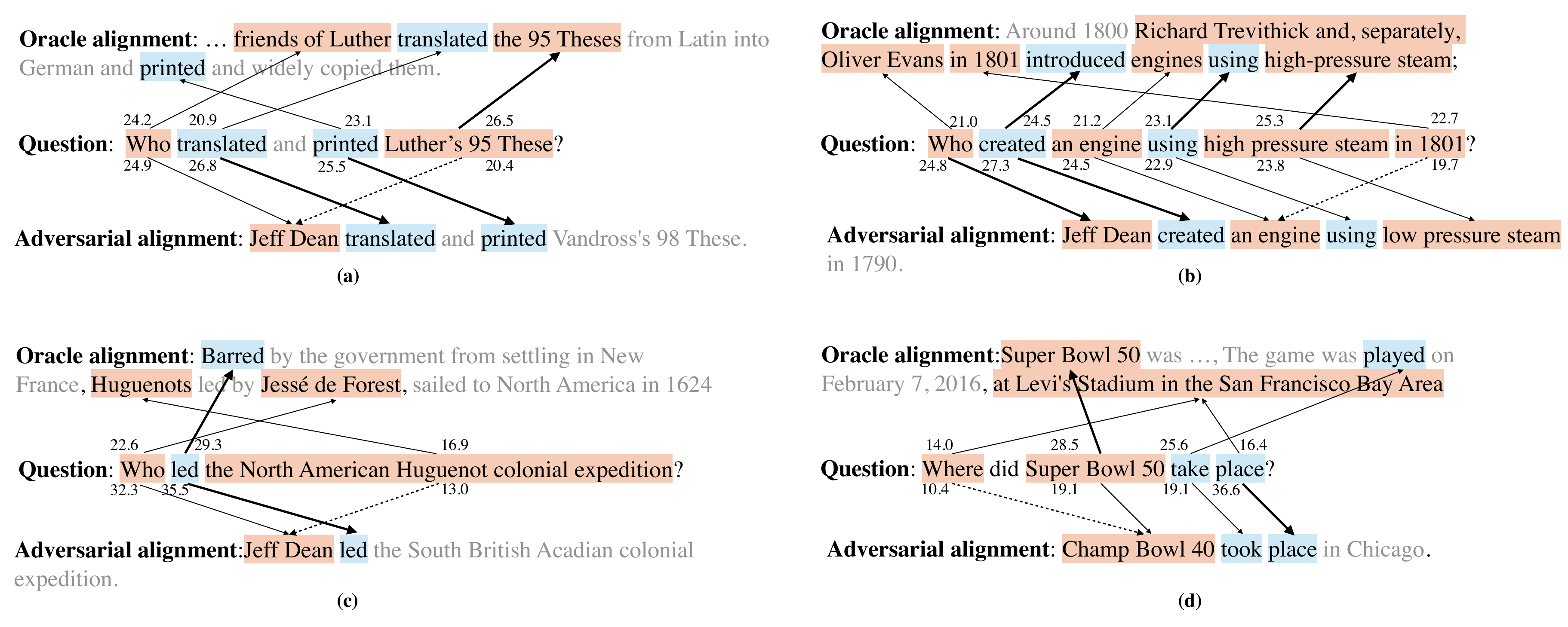}
\vspace{-0.8cm}
\caption{Examples of alignment of our model on \texttt{addOneSent}: both the correct alignment and also adversarial alignment are shown. The numbers are the actual alignment scores of the model's output. Dashed arrows denote the least reliable alignments and bolder arrows denote the alignment that contribute more to the model's prediction.}
\vspace{-0.2cm}
    \label{fig:case_study}
\end{figure*}




\paragraph{Comparison to BERT}
\newcite{desai2020calibration} show that pre-trained transformers like BERT are well-calibrated on a range of tasks. Since we are rejecting the unreliable predictions to improve the precision of our model, we reject the same number of examples for the baseline using the posterior probability of the BERT QA predictions. To be specific, we rank the predictions of all examples by the sum of \texttt{start} and \texttt{end} posterior probabilities and compute the F1 score on the top $k$ predictions.

\subsection{Results on Constrained Alignment}

\paragraph{On Adversarial SQuAD, the confidence scores of a normal BERT QA model do not align with its performance.} From Figure~\ref{fig:f1_coverage_curve}, we find that the highest-confidence answers from BERT (i.e., in low coverage settings) are very inaccurate. One possible explanation of this phenomenon is that BERT overfits to the pattern of lexical overlap, and is actually most confident on adversarial examples highly similar to the input. In general, BERT's confidence is not an effective heuristic for increasing accuracy.

\paragraph{Hard entity constraints improve the precision but are not flexible.} Figure~\ref{fig:f1_coverage_curve} also shows that by adding a hard entity constraint, we achieve a 71.4 F1 score which is an 8.6 improvement over the unconstrained model at a cost of only 60\% of samples being covered. Under the hard entity constraint, the model is not able to align to the nodes in the adversarial sentence, but the performance is still lower than what it achieves on normal SQuAD. We examine some of the error cases and find that for a certain number of samples, there is no path from the node satisfying the constraint to the node containing the answer (e.g. they hold a more complex discourse relation while we only consider coreference as cross-sentence relation). In such cases, our method cannot find the answer.

\paragraph{A smaller worst alignment gap indicates better performance.} As opposed to BERT, our alignment score is well calibrated on those adversarial examples. This substantiates our claim that those learned alignment scores are good indicators of how trustful alignment pairs are. Also, we see that when the coverage is the same as the entity constraint, the performance under the alignment score constraint is even better. The alignment constraints are simultaneously more flexible than the hard constraint and also more effective.  




\subsection{Case Study on Alignment Scores}
In this section, we give several examples of the alignment and demonstrate how those scores can act as an explanation to the model's behavior. Those examples are shown in Figure~\ref{fig:case_study}.

As shown by the dashed arrows, all adversarial alignments contain at least one alignment with significantly lower alignment score. The model is overconfident towards the other alignments with a high lexical overlap as shown by the bold arrows. These overconfident alignments also show that the predicate alignment learned on SQuAD-1.1 is not reliable. To further improve the quality of predicate alignment, either a more powerful training set or a new predicate alignment module is needed. 

Crucially, with these scores, it is easy for us to interpret our model's behavior. For instance, in example (a), the very confident predicate alignment forces \emph{Luther's 95 Theses} to have no choice but align to \emph{Jeff Dean}, which is unrelated. Because we have alignments over the sub-parts of a question, we can inspect our model's behavior in a way that the normal BERT QA model does not allow. We believe that this type of debuggability provides a path forward for building stronger QA systems in high-stakes settings.



\section{Related Work}

\paragraph{Adversarial Attacks in NLP.} Adversarial attacks in NLP may take the form of adding sentences like adversarial SQuAD \cite{jia2017adversarial}, universal adversarial triggers \cite{Wallace2019Triggers}, or sentence perturbations: \newcite{ribeiro2018semantically} propose deriving transformation rules, \newcite{ebrahimi2018hotflip} use character-level flips, and \newcite{iyyer2018adversarial} use controlled paraphrase generation. The highly structured nature of our approach makes it more robust to such attacks and provides hooks to constrain the system to improve performance further.

\paragraph{Neural module networks.} Neural module networks are a class of models that decompose a task into several sub-tasks, addressed by independent neural modules, which make the model more robust and interpretable~\cite{andreas2016neural, hu2017learning, cirik2018using, hudson2018compositional, jiang2019self}. Like these, our model is trained end-to-end, but our approach uses structured prediction and a static network structure rather than dynamically assembling a network on the fly. Our approach could be further improved by devising additional modules with distinct parameters, particularly if these are trained on other datasets to integrate additional semantic constraints.

\paragraph{Unanswerable questions} Our approach rejects some questions as unanswerable. This is similar to the idea of unanswerable questions in SQuAD 2.0~\cite{rajpurkar2018know}, which have been studied in other systems~\cite{hu2018read}. However, techniques to reject these questions differ substantially from ours -- many SQuAD 2.0 questions require not only a correct alignment between the question and context but also need to model the relationship between arguments, which is beyond the scope of this work and could be a promising future work. Also, the setting we consider here is more challenging, as we do not assume access to such questions at training time.

\paragraph{Graph-based QA}  \newcite{khashabi2018question} propose to answer questions through a similar graph alignment using a wide range of semantic abstractions of the text. Our model differs in two ways: (1) Our alignment model is trained end-to-end while their system mainly uses off-the-shelf natural language modules. (2) Our alignment is formed as node pair alignment rather than finding an optimal sub-graph, which is a much more constrained and less flexible formalism. \newcite{sachan2015learning,sachan2016machine} propose to use a latent alignment structure most similar to ours. However, our model supports a more flexible alignment procedure than theirs does, and can generalize to handle a wider range of questions and datasets.  

Past work has also decomposed complex questions to answer them more effectively \cite{talmor2018web,min2019multi,perez2020decompose}. \newcite{wolfson2020break} further introduce a Question Decomposition Meaning Representation (QDMR) to explicitly model this process. However, the questions they answer, such as those from HotpotQA \cite{yang2018hotpotqa}, are \emph{fundamentally} designed to be multi-part and so are easily decomposed, whereas the questions we consider are not. Our model theoretically could be extended to leverage these question decomposition forms as well.


\section{Discussion and Conclusion}

We note a few limitations and some possible future directions of our approach. First, errors from SRL and coreference resolution systems can propagate through our system. However, because our graph alignment is looser than those in past work, we did not observe this to be a major performance bottleneck. The main issue here is the inflexibility of the SRL spans. For example, not every SRL span in the question can be appropriately aligned to a single SRL span in the context. Future works focusing on the automatic span identification and alignment like recent work on end-to-end coreference systems~\cite{lee2017end}, would be promising.

Second, from the error analysis we see that our proposed model is good at performing noun phrase alignment but not predicate alignment, which calls attention to the better modeling of the predicate alignment process. For example, we can decompose the whole alignment procedure into separate noun phrase and predicate alignment modules, in which predicate alignment could be learned using different models or datasets.

Finally, because our BERT layer looks at the entire question and answer, our model can still leverage uninterpretable interactions in the text. We believe that modifying the training objective to more strictly enforce piecewise comparisons could improve interpretability further while maintaining strong performance.

In this work, we presented a model for question answering through sub-part alignment. By structuring our model around explicit alignment scoring, we show that our approach can generalize better to other domains. Having alignments also makes it possible to filter out bad model predictions (through score constraints) and interpret the model's behavior (by inspecting the scores).

\section*{Acknowledgments}

This work was partially supported by NSF Grant IIS-1814522 and NSF Grant SHF-1762299. The authors acknowledge the Texas Advanced Computing Center (TACC) at The University of Texas at Austin for providing HPC resources used to conduct this research. Results presented in this paper were obtained using the Chameleon testbed supported by the National Science Foundation.
Thanks as well to the anonymous reviewers for their helpful comments.

\bibliography{anthology,custom}
\bibliographystyle{acl_natbib}

\appendix

\section{Adversarial Datasets}

\paragraph{Added sentences}
\newcite{jia2017adversarial} propose to append an adversarial distracting sentence to the normal SQuAD development set to test the robustness of a QA model. In this paper, we use the two main test sets they introduced: \texttt{addSent} and \texttt{addOneSent}. Both of the two sets augment the normal test set with adversarial samples annotated by Turkers that are designed to look similar to question sentences. In this work, we mainly focus on the adversarial examples.

\paragraph{Universal Triggers}
\newcite{Wallace2019Triggers} use a gradient based method to find a short trigger sequence. When they insert the short sequence to the original text, it will trigger the target prediction in the sequence independent of the rest of the passage content or the exact nature of the question. For QA, they generate different triggers for different types of questions including ``who'', ``when'', ``where'' and ``why''.

\paragraph{Datasets from MRQA}
For Natural Questions~\cite{kwiatkowski2019natural}, NewsQA~\cite{trischler2017newsqa}, BioASQ~\cite{bioasq2015} and TextbookQA~\cite{kembhavi2017you}, we use the pre-processed datasets from MRQA~\cite{fisch2019mrqa}. They differ from the original datasets in that only the paragraph containing the answer is picked as the context and the maximum length of the context is cut to 800 tokens.

\section{Results on SQuAD addOneSent}
The results of our model compared to BERT QA on \texttt{SQuAD addOneSent} is shown in Table~\ref{tab:res_squad_adv_appendix}. Here we see the results on \texttt{addOneSent} and \texttt{addSent} generally have the same trend. The \textbf{global train+inf} helps more on the more difficult \texttt{addSent}.

\begin{table*}[t]
\small
\centering
\renewcommand{\tabcolsep}{1.3mm}
\begin{tabular}{ l | c c | c c | c c }
\toprule
            & \multicolumn{2}{c|}{SQuAD normal} & \multicolumn{2}{c|}{SQuAD addSent}& \multicolumn{2}{c}{SQuAD addOneSent} \\
\midrule
            & ans in wh & F1 & ans in wh & F1 & ans in wh & F1 \\
\midrule
Sub-part Alignment & 84.7 & 84.5 & 49.5 & \textbf{50.5}& 61.9 & \textbf{62.8} \\
- global train+inf  & 85.8  & 85.2  & 45.0  & 46.8 & 58.9 & 59.6 \\
- ans from full sent & 84.7  & 81.8  & 49.5  & 46.7 & 61.9 & 59.2 \\
\midrule
BERT QA & $-$ & 87.8 & $-$ & 39.2 & $-$ & 52.6 \\
\bottomrule
\end{tabular}
\caption{The performance and ablations of our proposed model on the development set of SQuAD normal, SQuAD addSent, and SQuAD addOneSent. $-$ \textbf{global train+inf} denotes the locally trained and evaluated model. $-$ \textbf{ans from full sent} denotes extracting the answer using only the wh-aligned node. \textbf{ans in wh} denotes the percentage of answers found in the span aligned to the wh-span, and F1 denotes the standard QA performance measure.}
\vspace{-0.5cm}
\label{tab:res_squad_adv_appendix}
\end{table*}

\end{document}